\documentclass[10pt,twocolumn,letterpaper]{article}

\usepackage{cvpr}              

\usepackage{adjustbox}
\usepackage{multicol}
\usepackage{multirow}
\usepackage{bbding}  
\usepackage{pifont}  
\usepackage{xcolor}
\usepackage{caption}  
\usepackage{makecell} 

\definecolor{cvprblue}{rgb}{0.21,0.49,0.74}
\usepackage[pagebackref,breaklinks,colorlinks,allcolors=cvprblue]{hyperref}

\def\paperID{12604} 
\def\confName{CVPR}
\def\confYear{2026}

\title{Bilingual Text-to-Motion Generation: A New Benchmark and Baselines}

\author{
Wanjiang Weng$^{1,2}$\footnote[1]{}
\quad \quad Xiaofeng Tan$^{1,2}$\footnote[1]{}
\quad Xiangbo Shu$^{3}$\\ 
\quad Guo-Sen Xie$^{3}$
\quad Pan Zhou$^{4}$
\quad Hongsong Wang$^{1,2}$\\ \\
$^{1}$School of Computer Science and Engineering, Southeast University, Nanjing 210096, China
\\ $^{2}$Key Laboratory of New Generation Artificial Intelligence Technology and  Its Interdisciplinary \\  Applications (Southeast University), Ministry of Education, China \\ \quad $^{3}$Nanjing University of Science and Technology  \\
 $^{4}$Singapore Management University \\
\small \texttt{\{wjweng, xiaofengtan, hongsongwang\}@seu.edu.cn}
}
\begin{document}

\maketitle

\renewcommand{\thefootnote}{\fnsymbol{footnote}} 
\footnotetext{%
\footnotemark[1] Equal Contribution.\\
}

\begin{abstract}
Text-to-motion generation holds significant potential for cross-linguistic applications, yet it is hindered by the lack of bilingual datasets and the poor cross-lingual semantic understanding of existing language models. To address these gaps, we introduce BiHumanML3D, the first bilingual text-to-motion benchmark, constructed via LLM-assisted annotation and rigorous manual correction. Furthermore, we propose a simple yet effective baseline, Bilingual Motion Diffusion (BiMD), featuring Cross-Lingual Alignment (CLA). CLA explicitly aligns semantic representations across languages, creating a robust conditional space that enables high-quality motion generation from bilingual inputs, including zero-shot code-switching scenarios. Extensive experiments demonstrate that BiMD with CLA achieves an FID of 0.045 vs. 0.169 and R@3 of 82.8\% vs. 80.8\%, significantly outperforms monolingual diffusion models and translation baselines on BiHumanML3D, underscoring the critical necessity and reliability of our dataset and the effectiveness of our alignment strategy for cross-lingual motion synthesis. The dataset and code are released at \href{https://wengwanjiang.github.io/BilingualT2M-page}{https://wengwanjiang.github.io/BilingualT2M-page}
\end{abstract}

\section{Introduction}\label{intro} 
\begin{figure}[tb]
    \centering
    \includegraphics[width=1\linewidth]{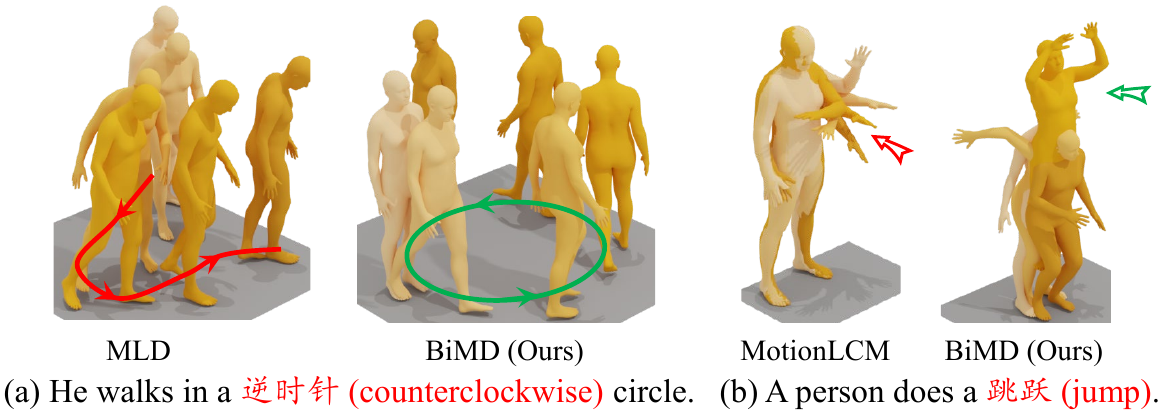}
    \caption{Code-switching in bilingual text-to-motion generation. Existing text-to-motion methods (e.g., MLD~\cite{Chen2023} and MotionLCM~\cite{Dai2025}) struggle to interpret mixed-language inputs, leading to incorrect motion semantics, while our BiMD generates semantically consistent motions.}
    \label{fig:codesw}
    \vspace{-5pt}
\end{figure}

Text-to-motion generation, which synthesizes 3D human motions from textual descriptions, has become increasingly important for gaming, filmmaking, and robotics~\cite{Chen2023, Dai2025, Guo2022, jiang2024motiongpt,motionstreamer,jeong2025hgm}. As these applications expand to global and multicultural markets, the ability to generate motion from non-English text is essential for broader accessibility and cultural fidelity. However, current text-to-motion methods operate almost exclusively in English, limiting their applicability to non-English-speaking users and failing to capture culturally specific human actions. To advance bilingual text-to-motion generation, three key challenges must be addressed.

\begin{figure*}[t]
    \centering
    \includegraphics[width=\linewidth]{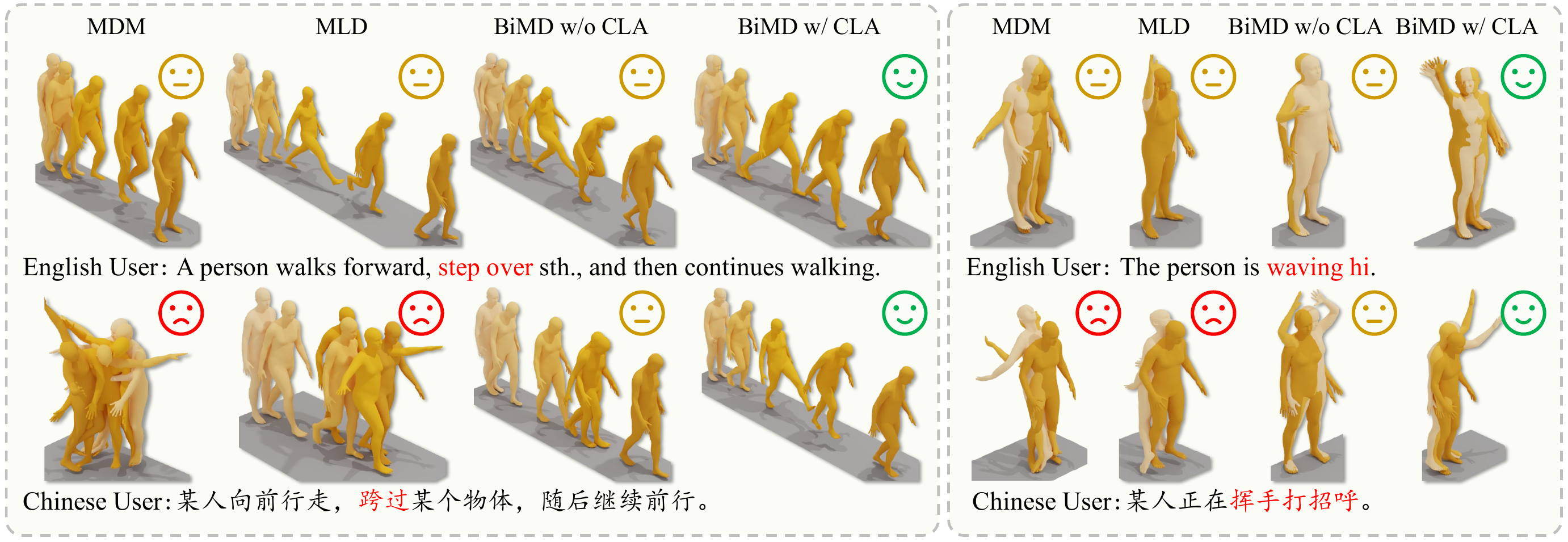} 
    \caption{Visual comparison of bilingual text-to-motion generation. Existing methods such as MDM~\cite{Tevet2023} and MLD~\cite{Chen2023} exhibit limitations in bilingual processing. Directly using pretrained multilingual encoders results in imbalanced performance across languages. Our BiMD with Cross-Lingual Alignment (CLA) generates accurate motions from both English and Chinese descriptions.}
    \label{fig:intro}
\end{figure*}

Firstly, the scarcity of non-English and bilingual text-to-motion datasets remains a critical bottleneck. While several English text-to-motion datasets exist~\cite{Guo2022, kit, babel, finemotion}, there are no publicly available, high-quality motion datasets annotated in other major languages. We focus on the English-Chinese language pair as a representative case study, given the substantial linguistic and cultural disparity between the two languages, and its capacity to serve as a testbed for cross-lingual challenges across diverse language families~\cite{taiyidiff, seedream2, li2024hunyuan}.

Second, a translate-then-generate pipeline cannot serve as a viable alternative to a dedicated bilingual model. Such a pipeline must commit to a single target language, which makes it structurally unable to process code-switching inputs. Code-switching, where users mix two languages within a single utterance, is the natural way for bilingual speakers to communicate, and even simple cases pose a fundamental challenge for existing methods. As shown in Figure~\ref{fig:codesw}, when key action words are code-switched between English and Chinese (e.g., \textit{He walks in a NiShiZhen circle''} or \textit{A person does a TiaoYue''}), existing methods such as MLD and MotionLCM fail to correctly interpret the mixed-language inputs, leading to incorrect motion semantics. Beyond these straightforward cases, the challenge becomes even more acute for culturally specific actions that lack cross-lingual equivalents~\cite{kabra-etal-2023-multi}. Movements such as ZuoYi'' (a formalized bow with folded hands) or TanglangQuan'' (mantis fist, a martial arts stance) carry kinematic details and cultural connotations that are entirely absent from the target language vocabulary, rendering translation-based approaches fundamentally inadequate. A dedicated bilingual model is therefore not merely preferable but necessary.


Third, although existing multilingual text encoders and pretrained models perform well across various NLP and vision-language domains, they are not well suited for motion generation~\cite{altclip, conneau2019unsupervised, multilingualBERT}. These general-purpose models are pretrained on corpora with highly imbalanced language distributions and lack motion-specific semantics~\cite{yang2024language}. As a result, they exhibit significant performance gaps across languages, restricting cross-lingual generalization and code-switching capability. As shown in Figure~\ref{fig:intro}, when conditioned on semantically equivalent English and Chinese inputs, existing methods fail to capture key motion details such as stepping over an object or waving. Directly adopting these general-purpose encoders as text conditions compromises both semantic accuracy and cross-lingual consistency of the generated motions.

While research in bilingual generative models is gaining traction, most efforts remain domain-specific. For instance, SOKE~\cite{zuo2024signs} develops a bilingual 3D sign language avatar generator, while SEEDREAM2~\cite{seedream2} and Hunyuan-DiT~\cite{li2024hunyuan} introduce bilingual image diffusion transformers. However, sign language generation is restricted to upper-body symbolic gestures without modeling full-body dynamics, and image diffusion operates on spatial pixel distributions without capturing the temporal joint-level kinematics required for motion synthesis. These fundamental differences leave bilingual text-to-motion generation an open challenge. To the best of our knowledge, this is the first dedicated effort to explore bilingual text-to-motion generation.

\paragraph{Contributions:} Our first contribution is the introduction of BiHumanML3D, the first bilingual text-to-motion benchmark, along with a corresponding baseline model, Bilingual Motion Diffusion (BiMD). To address the scarcity of bilingual text-motion datasets, we extend the widely used HumanML3D~\cite{Guo2022} by constructing its bilingual version through a multi-stage annotation pipeline combining large language models and manual correction. This pipeline is language-general and can be extended to other language pairs.

Second, to mitigate the performance imbalance between languages, we propose Cross-Lingual Alignment (CLA), a module that explicitly aligns semantic representations across languages into a unified conditional space. During training, BiMD is conditioned on consistent cross-lingual semantic representations, enabling balanced bilingual text-to-motion generation. We employ a lightweight adapter to refine fine-grained semantic alignment, yielding more effective bilingual conditioning.

Finally, comprehensive experiments demonstrate that BiMD with CLA achieves an FID of 0.045 (vs. 0.169 without CLA) and R@3 of 82.8\% (vs. 80.8\%), outperforming monolingual and direct-translation baselines. BiMD also demonstrates strong zero-shot cross-lingual transfer and robust code-switching performance, validating the necessity of bilingual training and the effectiveness of CLA. As the first bilingual text-to-motion benchmark, we believe BiHumanML3D will offer valuable insights for the community and facilitate broader multilingual motion generation research.

\section{Related Works}\label{rw}

\subsection{Text-to-Motion Generation} 
Text-to-motion generation represents a critical task in computer vision, exhibiting rapid advancements in recent years~\cite{zhang2025large, zhang2023finemogen, zhang2023remodiffuse,zhong2023attt2m, motionclip,yuan2025mogents}. Specifically, Tevet et al.\cite{Tevet2023} and Zhang et al.\cite{Zhang2024} first propose diffusion models to address text-driven motion generation, laying the groundwork for the following innovations. Subsequently, Dai et al.~\cite{Dai2025} present MotionLCM, a real-time controllable model that refines motion-latent diffusion, enabling precise spatiotemporal control via few-step inference. Zhang et al.\cite{motionmaba} introduce motion mamba, a state-space framework that leverages hierarchical temporal and bidirectional spatial modules to improve efficiency and long-sequence modeling. Xiao et al.~\cite{motionstreamer} introduce MotionStreamer, a probabilistic autoregressive framework that utilizes a continuous causal latent space to achieve accurate, streaming motion generation. However, these works are still confined to monolingual settings. Bilingual text-to-motion generation remains a challenging area due to the lack of datasets and misalignment between different languages.

\subsection{Multilingual Generation} 
Multilingual generative modeling has recently garnered significant attention in the text, vision, and audio domains, typically leveraging cross-lingual encoders or specialized language adapters to bridge linguistic gaps \cite{seedream2, taiyidiff, li2024hunyuan, qin2024multilingual}. In the text-generation domain, recent benchmarks such as MultiSocial advance multilingual machine-generated-text detection for social-media posts, further validating cross-lingual knowledge transfer~\cite{macko2025multisocial}. For vision–language tasks, multilingual diffusion systems like TextGen utilize lightweight adapters and cross-modal encoders to enable high-quality text-to-image generation across over a hundred languages~\cite{zhang2024control}. However, research on multilingual motion generation remains limited. Recent approaches like SOKE~\cite{zuo2024signs} leverage pretrained language models to generate sign language motions, though current methods are restricted to upper-body synthesis without full-body movement generation. In this work, we present the first exploration of multilingual settings for text-to-motion generation. Specifically, we focus on a simplified yet fundamental bilingual scenario (English–Chinese) due to data scarcity and alignment challenges inherent to multilingual contexts. To this end, we introduce a pioneering bilingual benchmark and propose effective baselines tailored explicitly for bilingual text-to-motion generation, including zero-shot generalization and code-switching support, paving the way for future research in broader multilingual motion generation.

\begin{figure}[tb]
    \centering
    \includegraphics[width=1\linewidth]{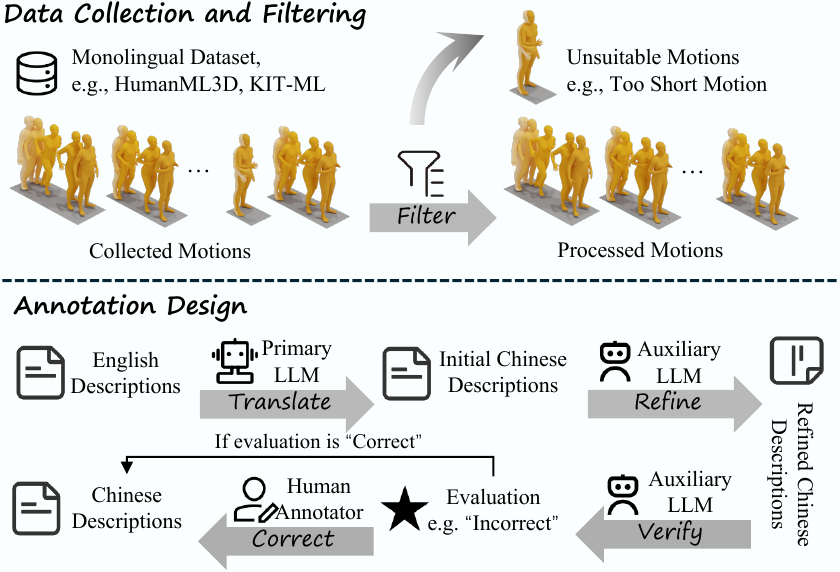}
    \caption{Pipeline for constructing our bilingual HumanML3D dataset. The data collection and filtering process removes unsuitable motions, ensuring high-quality motion-text pairs for annotation. The annotation pipeline begins with an initial translation stage, followed by a refinement stage to address translation issues. Finally, human annotators manually verify and correct the translation with LLM, ensuring linguistic and contextual accuracy.}
    \label{fig:build_bih3d}
    \vspace{-10pt}
\end{figure}
\section{Bilingual HumanML3D Dataset}\label{datasets} 
Despite significant progress in text-to-motion synthesis, their reliance on English-only datasets limits their usability in multilingual applications, thereby reducing accessibility for non-English speakers and restricting cultural diversity in global industries such as animation and robotics. 

A straightforward approach to multilingual motion synthesis might involve translating non-English text into English using off-the-shelf translation tools. However, we observe that such direct translation often fails to capture subtle motion-related semantics, resulting in a mismatch between text and motions. 
Thanks to the powerful Large Language Models (LLMs), we propose an elaborate multi-stage LLM-assisted annotation pipeline to construct a high-quality bilingual motion dataset, enabling bilingual motion synthesis via three stages, as shown in Figure~\ref{fig:build_bih3d}. Constructing details are described below.

\subsection{Data Collection and Filtering} 
Since English text-to-motion datasets are well-established and widely used, we extend HumanML3D~\cite{Guo2022}, a widely recognized dataset, into a bilingual version. Following prior studies \cite{Chen2023, Tevet2023}, we first filter out motions that are too short or too long, as they often do not provide meaningful and reliable training signals. This ensures that only high-quality motion-caption pairs are retained for subsequent annotation.  Specifically, we remove 1,058, 186, and 60 motions from the training, testing, and validation sets, respectively, accounting for $4.5\%$, $4.2\%$, and $4.1\%$ of their corresponding datasets in total.

\subsection{Annotation Design} 
We build a robust LLM-assisted pipeline to extend HumanML3D with bilingual (English–Chinese) motion descriptions while preserving motion semantics and ensuring high-quality annotations. As illustrated in Figure~\ref{fig:build_bih3d}, the pipeline consists of three stages:
(1) An initial LLM (e.g., DeepSeek~\cite{liu2024deepseek}) translates the English motion descriptions into the target language.
(2) A refinement model (e.g., Qwen~\cite{yang2024qwen2}) corrects translation artifacts such as gender bias, literal translation, and unnatural phrasing.
(3) Human annotators evaluate translation quality with the assistance of an LLM, where entries flagged as uncertain or incorrect are manually reviewed to ensure linguistic and contextual fidelity. We additionally used rule-based checks to enforce linguistic consistency.

Since HumanML3D and KIT-ML adopt incompatible motion representations, we use only HumanML3D motions for bilingual extension, while providing additional Chinese descriptions for both datasets to enhance language diversity and improve bilingual alignment.

The pipeline processed approximately 35k motion descriptions. The two LLM-based stages achieved success rates of 97\% and 66\%, respectively. To ensure annotation reliability, a team of 17 annotators manually verified 53\% of all entries (covering approximately 18.5k descriptions across HumanML3D and KIT-ML), including all samples containing more than six verbs, which are prioritized due to their higher semantic complexity. We further employed rule-based consistency checks (e.g., directional terms such as front/back/left/right) to detect systematic errors. On average, each annotator processed $\sim$1,100 descriptions with a measured inter-annotator agreement rate of 91.3\% on a randomly sampled subset of 500 overlapping entries.

The resulting Bilingual HumanML3D dataset contains 13,312 motions with high-quality bilingual annotations, enabling multilingual text-to-motion generation and improving the cross-lingual adaptability of motion models.

\section{Bilingual Text-to-Motion Generation}\label{methods}
 \begin{figure*}[tbp]
   \centering
   \includegraphics[width=1\textwidth]{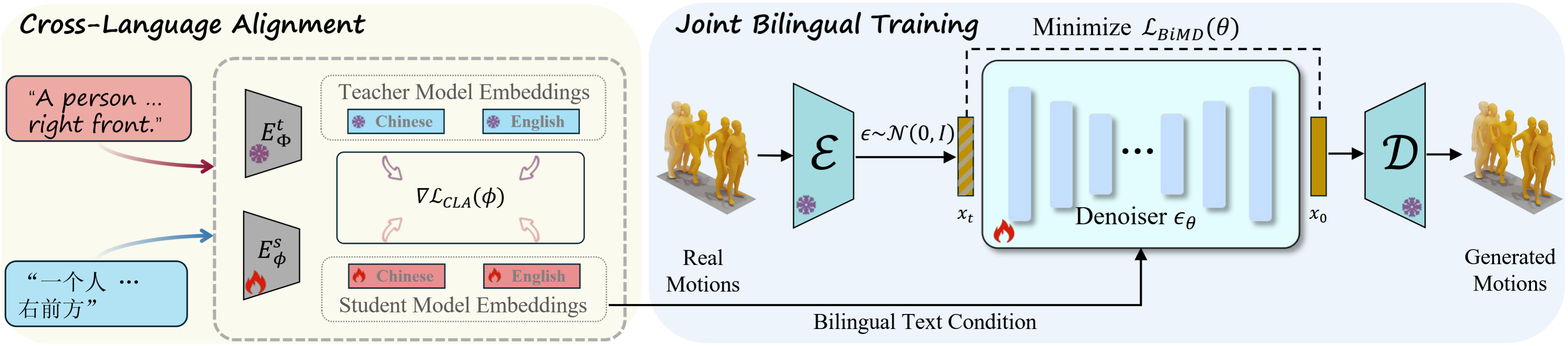}
   \caption{Framework for training the bilingual motion diffusion model. We align English and Chinese text embeddings in a shared latent space by freezing the teacher model $E^t_{\Phi}$ and fine-tuning the student model $E^s_{\phi}$ with the cross-lingual alignment loss $\mathcal{L}_{CLA}$ in  Eq.~\eqref{eq:bikl}. The aligned student model $E^s_{\phi}$ then provides text conditions for training the diffusion model $\epsilon_\theta$, enabling bilingual motion generation while minimizing $\mathcal{L}_{\text{BiMD}}$ in Eq.~\eqref{eq:mld_loss}.}
   \label{fig:fig_main_1}
\end{figure*}

With the introduction of our bilingual dataset, a straightforward approach would be to train separate models for each language. However, this strategy is not only computationally expensive but also fails to leverage the underlying semantic similarities between languages. 
To address this, we propose a unified Bilingual Motion Diffusion Model (BiMD) that enhances motion diversity and adaptability across languages. 
As illustrated in Figure~\ref{fig:fig_main_1}, our approach incorporates two key components: cross-lingual alignment (CLA), which explicitly minimizes the distance between English and Chinese text representations in a shared latent space, and bilingual diffusion training, which conditions the denoiser on cross-lingually aligned embeddings. Together, these components ensure high-quality motion generation from text descriptions in multiple languages.

\subsection{Cross-Lingual Alignment} 
To ensure consistent conditional representations for motion descriptions across languages, we align text embeddings to enable the text encoder to process bilingual inputs interchangeably. Instead of directly using off-the-shelf multilingual encoders (e.g., T5 or mBERT~\cite{2020t5, multilingualBERT}), we employ fine-tuning to enhance cross-lingual generalization and zero-shot capabilities, addressing the limitations of generic pretrained models. Inspired by AltCLIP~\cite{altclip}, we fine-tune a pretrained multilingual model $E^s_\phi$ via knowledge distillation (KD), leveraging a teacher model's robust embeddings (e.g., OpenCLIP~\cite{cherti2023reproducible}) while adapting to motion-specific semantics. This approach avoids the computational cost and inefficiency of training from scratch for specialized tasks. Specifically, we align the text embeddings of the student model $E^s_\phi$, e.g., XLM~\cite{conneau2019unsupervised}, with those of a teacher model $E^t_\Phi$, e.g., OpenCLIP~\cite{cherti2023reproducible} by optimizing:
\begin{equation}
	\label{eq:bikl}
	\mathcal{L}_{\mathrm{CLA}} (\phi) = \mathrm{D}_\mathrm{KL}(\mathrm{F}^{\Phi}_{\text{en}} \mid\mid \mathrm{F}^{\phi}_{\text{lang}}) + 
	\mathrm{D}_\mathrm{KL}(\mathrm{F}^{\phi}_{\text{lang}} \mid\mid \mathrm{F}^{\Phi}_{\text{en}}),
\end{equation}  
where $\mathrm{F}^{\Phi}_{\text{en}}$ denotes the English text embeddings from the \emph{frozen} teacher $E^t_\Phi$, and $\mathrm{F}^{\phi}_{\text{lang}}$ denotes either English or Chinese embeddings from the \emph{trainable} student $E^s_\phi$. To apply $\mathrm{D}_\mathrm{KL}$ to embedding vectors, we convert them into probability distributions via softmax normalization with a temperature $\tau{=}0.05$ over the embedding dimensions. We adopt a symmetric KL divergence so that the student is trained both to match the teacher's distribution and to sharpen its own.  

Minimizing $\mathcal{L}_{CLA}$ ensures that $E^s_\phi$ produces interchangeable embeddings for both languages, preserving motion fidelity without separate per-language models.

\paragraph{Teacher--Student Model Selection:}
We select OpenCLIP~\cite{cherti2023reproducible} as the teacher for its strong English text--vision alignment, and XLM-Base~\cite{conneau2019unsupervised} as the student for its inherent multilingual capacity across 100 languages. Although OpenCLIP is predominantly English-centric, this is precisely its advantage as a teacher: it provides high-quality, well-structured English embeddings as the alignment target. The student (XLM) then learns to project Chinese embeddings into this well-organized space. We note that replacing the teacher with a multilingual CLIP variant (e.g., Chinese-CLIP~\cite{chinese-clip}) yields a 2.1\% drop in R@3, as the multilingual teacher's Chinese embeddings are less well-structured than OpenCLIP's English space, confirming the soundness of our asymmetric design choice.

\subsection{Bilingual Motion Diffusion Model} 
Existing diffusion-based motion generation methods consist of a forward process and reverse process~\cite{ho2020denoising,Tevet2023}. The forward process gradually adds noise into the real motion $\mathbf{x}_0$ over $T$ steps, forming a Markovian process:
\begin{equation}
    q(\mathbf{x}_t | \mathbf{x}_{t-1}) = \mathcal{N}(\mathbf{x}_t; \sqrt{1 - \beta_t} \mathbf{x}_{t-1}, \beta_t \mathbf{I}),
    \label{eq:forward}
\end{equation}
where $\beta_t \in (0,1)$ controls the noise schedule. As $T$ increases, $\mathbf{x}_T$ approaches a standard Gaussian distribution.

The reverse process generates motion from a Gaussian noise $\mathbf{x}_t \sim \mathcal{N}(0, I)$ by iterative denoising:
\begin{equation}
    p_\theta(\mathbf{x}_{t-1} | \mathbf{x}_t) = \mathcal{N}(\mathbf{x}_{t-1}; \mu_\theta(\mathbf{x}_t, t), \Sigma(\mathbf{x}_t, t)),
    \label{eq:reverse}
\end{equation}
where $\theta$ denotes learnable parameters. Then we reparameterize $\mu_\theta$ to predict the noise $\epsilon_\theta$:
\begin{equation}
    \mu_\theta(\mathbf{x}_t, t) = \frac{1}{\sqrt{\alpha_t}} \left( \mathbf{x}_t - \frac{\beta_t}{\sqrt{1 - \bar{\alpha}_t}} \epsilon_\theta(\mathbf{x}_t, t) \right),
\end{equation}
where $\alpha_t = 1 - \beta_t$, $\bar{\alpha}_t = \prod_{s=1}^t \alpha_s$. This reduces the training objective to:
\begin{equation}
    \mathcal{L}_{\text{DM}}(\theta) = \mathbb{E}_{\mathbf{x}_0, t, \epsilon \sim \mathcal{N}(0,I)} \left[ \| \epsilon - \epsilon_\theta(\mathbf{x}_t, t) \|^2 \right].
\end{equation}

However, diffusion on raw motion sequences is computationally intensive and sensitive to noise in motion capture results. Thus, we adopt a latent diffusion framework based on MLD~\cite{Chen2023}, enhancing efficiency and robustness by first compressing the raw motion $\mathbf{x}$ into a compact latent space $\mathbf{z}$ via a VAE \cite{vae}.

\subsection{Bilingual Diffusion Training} 
To enable robust bilingual motion generation, we enhance the motion diffusion model to produce linguistically invariant motions from bilingual text inputs. We build our model upon the motion latent diffusion model introduced by MLD~\cite{Chen2023}, conditioning it on cross-lingually aligned text embeddings, achieving fine-grained semantic correspondence across languages for effective bilingual conditioning. Specifically, instead of training two distinct models, we introduce a bilingual diffusion training strategy: each motion is randomly paired with either an English or a Chinese text description. This encourages the model to capture shared motion patterns while remaining sensitive to language-specific nuances. Our training loss is as follows:
\begin{equation}
	\label{eq:mld_loss}
	\mathcal{L}_{\text{BiMD}} (\theta) = \mathbb{E}_{\epsilon,t,c} \left[\left\| 
	\epsilon - \epsilon_\theta \left(\mathbf{z}_t,t, E_{\phi}^{s}\left(c_{s}\right)\right)
	\right\|^{2}_{2}\right],
\end{equation}  
where $c_s$ is the randomly selected English or Chinese description of motion latent $\mathbf{z}$, and $E_{\phi}^{s}(c_{s})$ represents its cross-lingually aligned text embedding. This approach enables the model to synthesize high-quality motion from bilingual text inputs without requiring separate models.  

Furthermore, we adopt classifier-free guidance (CFG)~\cite{ho2022classifier} to improve generation quality. During \emph{training}, the text condition is randomly dropped with probability $p{=}0.1$, so that the denoiser $\epsilon_\theta$ learns both conditional and unconditional distributions. At \emph{inference}, the two predictions are linearly combined:
\begin{equation}
    \label{cfg}
    \epsilon_\theta^s = s \epsilon_\theta ({z}_t, t, c_{s}) + (1 - s) \epsilon_\theta ({z}_t, t, \emptyset),
\end{equation}
where $s$ is the guidance scale ($s{=}7.5$ in all experiments); $s > 1$ amplifies the guidance effect. After the conditional denoising reverse process, the motion is efficiently reconstructed from the predicted $\mathbf{z}$ by a VAE decoder $\mathcal{D}$.

By integrating cross-lingual alignment with joint bilingual training, our method overcomes the English-only limitation of existing text-to-motion models, making text-driven motion generation more accessible and effective across languages. This paves the way for broader applications like virtual animation and multilingual interactions.

\section{Experiments}\label{exp}
\begin{table*}[ht]
    \centering
    \setlength{\tabcolsep}{5pt}
    \adjustbox{max width=0.95\linewidth}{
    \begin{tabular}{l l l l l l l l}
      \toprule
      \multirow{2}{*}{Method} & 
      \multirow{2}{*}{Publisher} & 
      \multicolumn{3}{c}{R Precision $\uparrow$} & 
      \multirow{2}{*}{FID $\downarrow$} & 
      \multirow{2}{*}{MM Dist $\downarrow$} & 
      \multirow{2}{*}{Diversity $\rightarrow$} \\
      \cmidrule(lr){3-5}
      ~ & ~ & Top 1 & Top 2 & Top 3 & & & \\
      \midrule
      Real       & - & 0.511 & 0.703 & 0.797 & 0.002 & 2.974 & 9.503 \\
      \midrule
      T2M \cite{Guo2022}   & CVPR'22 & 0.455$^{\pm 0.002}$ & 0.636$^{\pm 0.003}$ & 0.736$^{\pm 0.003}$ & 1.087$^{\pm 0.002}$ & 3.347$^{\pm 0.008}$ & 9.175$^{\pm 0.002}$ \\
      TEMOS \cite{petrovich2022temos} & ECCV'22 & 0.424$^{\pm 0.002}$ & 0.612$^{\pm 0.002}$ & 0.722$^{\pm 0.002}$ & 3.734$^{\pm 0.028}$ & 3.703$^{\pm 0.008}$ & 8.973$^{\pm 0.071}$ \\
      MDM \cite{Tevet2023} & NeurIPS'23 & 0.455$^{\pm 0.006}$ & 0.645$^{\pm 0.007}$ & 0.749$^{\pm 0.006}$ & 0.489$^{\pm 0.047}$ & 3.330$^{\pm 0.25}$ & 9.920$^{\pm 0.083}$ \\
      T2M-GPT \cite{Zhang_2023_T2M_GPT} & CVPR'23 & 0.492$^{\pm 0.003}$ & 0.679$^{\pm 0.002}$ & 0.775$^{\pm 0.002}$ & 0.141$^{\pm 0.005}$ & 3.121$^{\pm 0.009}$ & 9.722$^{\pm 0.082}$\\
      MLD \cite{Chen2023}     & CVPR'23 & 0.481$^{\pm 0.003}$ & 0.673$^{\pm 0.003}$ & 0.772$^{\pm 0.002}$ & 0.473$^{\pm 0.013}$ & 3.196$^{\pm 0.010}$ & 9.724$^{\pm 0.082}$ \\
      Mo.Diffuse \cite{Zhang2024} & TPAMI'24 & 0.491$^{\pm 0.001}$ & 0.681$^{\pm 0.001}$ & 0.775$^{\pm 0.001}$ & 0.630$^{\pm 0.001}$ & 3.113$^{\pm 0.001}$ & 9.410$^{\pm 0.049}$ \\
      OMG \cite{liang2024omg}  & CVPR'24 & - & - & 0.784$^{\pm 0.002}$   & 0.381$^{\pm 0.008}$ & - & 9.657$^{\pm 0.085}$ \\ 
      MoMask \cite{guo2024momask}  & CVPR'24 & 0.521$^{\pm 0.002}$ & 0.713$^{\pm 0.002}$ & 0.807$^{\pm 0.002}$ & \textbf{0.045}$^{\pm 0.008}$ & 2.958$^{\pm 0.008}$ & - \\
      MotionLCM \cite{Dai2025}  & ECCV'24 & 0.502$^{\pm 0.003}$ & 0.698$^{\pm 0.002}$ & 0.798$^{\pm 0.002}$ & 0.304$^{\pm 0.012}$ & 3.012$^{\pm 0.007}$ & 9.607$^{\pm 0.066}$ \\
      LMM-T$^2$ \cite{Zhang2024lmm}  & ECCV'24 & 0.496$^{\pm 0.002}$ & 0.685$^{\pm 0.002}$ & 0.785$^{\pm 0.002}$ & 0.415$^{\pm 0.002}$ & 3.087$^{\pm 0.012}$  & 9.176$^{\pm 0.074}$ \\ 
      Mo.Mamba \cite{Zhang2025}& ECCV'24 & 0.502$^{\pm 0.003}$ & 0.693$^{\pm 0.002}$ & 0.792$^{\pm 0.002}$ & 0.281$^{\pm 0.011}$ & 3.060$^{\pm 0.000}$  & 9.871$^{\pm 0.084}$\\ 
      ParCo \cite{Zou2025ParCo}  & ECCV'24 & 0.515$^{\pm 0.003}$ & 0.706$^{\pm 0.003}$ & 0.801$^{\pm 0.002}$ & 0.109$^{\pm 0.005}$ & 2.927$^{\pm 0.008}$ &9.576$^{\pm 0.088}$ \\

     MARDM \cite{meng2024rethinking}  & CVPR'25 & 0.500$^{\pm 0.004}$ & 0.695$^{\pm 0.003}$ & 0.795$^{\pm 0.003}$ & 0.114$^{\pm 0.007}$ & - & - \\
      MG-MotionLLM  \cite{wu2025mg} & CVPR'25 & 0.516$^{\pm 0.002}$ & 0.706$^{\pm 0.002}$ & 0.802$^{\pm 0.003}$ & 0.303$^{\pm 0.010}$ & 2.952$^{\pm 0.009}$ & 9.960$^{\pm 0.073}$ \\ 
      EnergyMoGen  \cite{zhang2025energymogen} & CVPR'25 & 0.526$^{\pm 0.003}$ & 0.718$^{\pm 0.003}$ & 0.815$^{\pm 0.002}$ & 0.176$^{\pm 0.006}$ & 2.931$^{\pm 0.007}$ &  \textbf{9.500}$^{\pm 0.091}$ \\ 
      InfiniteDreamer$^3$  \cite{infdreamer} & ICCV'25 & - & - & 0.679$^{\pm 0.007}$ & 0.047$^{\pm 0.006}$ & 3.150$^{\pm 0.003}$ &  9.580$^{\pm 0.087}$ \\ 
      GenM$^3$  \cite{genm3} & ICCV'25 & 0.511$^{\pm 0.003}$ & 0.705$^{\pm 0.002}$ & 0.804$^{\pm 0.002}$ & 0.046$^{\pm 0.002}$ & 2.852$^{\pm 0.009}$ &  9.675$^{\pm 0.087}$ \\ 
      \midrule
      BiMD w/o CLA (Ours)    & This work & 0.528$^{\pm 0.003}$ & 0.709$^{\pm 0.003}$ & 0.804$^{\pm 0.002}$ & 0.169$^{\pm 0.006}$ & 3.152$^{\pm 0.008}$ & 9.881$^{\pm 0.076}$ \\ 
      BiMD w/ CLA (Ours)    & This work & \textbf{0.558}$^{\pm 0.002}$ & \textbf{0.731}$^{\pm 0.003}$ & \textbf{0.828}$^{\pm 0.003}$ & \textbf{0.045}$^{\pm 0.003}$ & \textbf{2.803}$^{\pm 0.007}$ & \text{9.619}$^{\pm 0.085}$ \\ 
      \bottomrule
\end{tabular}
}
\caption{{Comparison of text-to-motion generation performance on the HumanML3D dataset.} These metrics are evaluated by the evaluator from TM2T~\cite{Guo2022}. The arrows $\uparrow$, $\downarrow$, and $\rightarrow$ indicate higher, lower, and closer-to-real-motion values are better, respectively. \textbf{Bold} highlights the best results.}
\label{tab:sota_humanmld3d}
 \vspace{-10pt}
\end{table*}
\subsection{Experimental Settings}
\paragraph{Datasets:} HumanML3D~\cite{Guo2022} datasets comprise 14,616 motion sequences, each paired with one or more text descriptions, resulting in a total of 44,970 annotations. The KIT-ML~\cite{kit} dataset comprises 3,911 motion sequences, each accompanied by 6,278 corresponding text descriptions. The proposed BiHumanML3D dataset extends the HumanML3D dataset into a bilingual dataset through an elaborate LLM-assisted annotations pipeline, incorporating bilingual text descriptions while preserving the original motion sequences. 

\paragraph{Implementation Details:} To achieve effective cross-lingual alignment, we employ OpenCLIP~\cite{cherti2023reproducible} as the teacher model and adopt XLM-Base~\cite{conneau2019unsupervised} as the backbone of the student model, optimized using the alignment loss $\mathcal{L}_{CLA}$ defined in Eq.(\ref{eq:bikl}). The student model is augmented by adding a linear projection layer to extract refined sentence embeddings, further facilitating cross-lingual semantic alignment. During optimization, we use the AdamW optimizer\cite{adamw} with a cosine decay learning rate schedule, preceded by a 500-step linear warm-up phase. Training is performed for 50 epochs with a batch size of 512 and an initial learning rate of 1e-4. Our bilingual motion diffusion model is based on MLD~\cite{Chen2023}, with modifications including the removal of the activation function after the text encoder and the addition of a linear projection layer after the VAE encoder.  

\paragraph{Evaluation Metrics:} Our experimental results are evaluated based on two key aspects: generation quality and alignment quality. For generation quality, we use Fr$\acute{\text{e}}$chet Inception Distance (FID) to measure the distributional difference between high-level features of generated and real motions, and Diversity to assess motion diversity by calculating variation among generated motions~\cite{fid}. For alignment quality, R-Precision evaluates motion-retrieval precision, assessing matching quality between generated motions and text descriptions, while Multi-Modal Distance (MM Dist) quantifies the distance between motions and their corresponding text descriptions. 

\subsection{Evaluation of Text-to-Motion Generation} 
To validate the necessity of bilingual diffusion models and the effectiveness of our cross-lingual alignment, we conducted extensive text-to-motion experiments on both the HumanML3D dataset and its bilingual extension, covering diverse training and evaluation scenarios.

\paragraph{Monolingual Text-to-Motion Generation:} Table \ref{tab:sota_humanmld3d} reports the results on the monolingual HumanML3D dataset. The results demonstrate that our model achieves the best performance through two main improvements. First, the CLA effectively enhances the capability of the text encoder. Compared to the BiMD without CLA, our approach improves the text-to-motion semantic alignment metric R Precision@1 by 5\%, and reduces the motion quality metric FID from 0.169 to 0.045, indicating significant gains in both alignment and generation quality. Second, compared to the backbone MLD~\cite{Chen2023}, our architectural modifications (removing the activation function after the text encoder and adding a linear projection layer after the VAE encoder) improve the compatibility between the text embeddings and the latent diffusion model, resulting in better latent representations for motion synthesis. With these improvements, our BiMD achieves state-of-the-art results across all major evaluation metrics.

\begin{figure}[tb]
    \centering
    \includegraphics[width=1\linewidth]{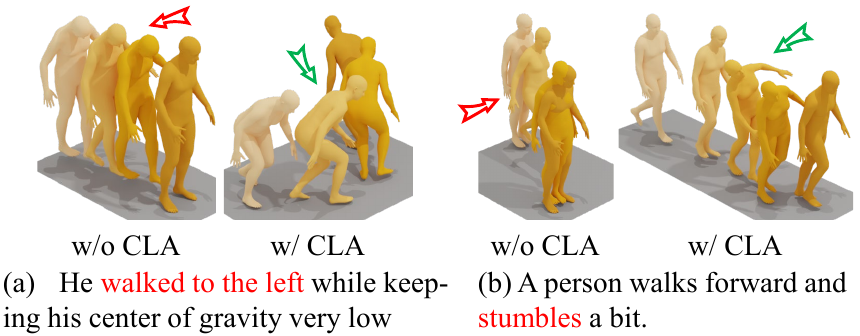}
    \caption{Visual comparison of zero‑shot cross‑lingual motion generation from Chinese‑only training to unseen English texts. We find that BiMD with CLA generates accurate, semantically aligned motions despite never having seen English during training.} 
    \label{fig:zero-shot}
    \vspace{-5pt}
\end{figure}

\paragraph{Bilingual Text-to-Motion Generation:} Table \ref{tab:bih3d_result} presents the results for bilingual text to motion generation on the BiHumanML3D dataset. Given the absence of bilingual motion generation methods in current literature, we retrain the diffusion models Mo.Diffuse, MLD, and MDM on BiHumanML3D and replace their original English text encoders with a multilingual text encoder XLM~\cite{conneau2019unsupervised} for a fair comparison. We then substitute XLM with our CLA encoder to evaluate the benefit of explicit cross-language semantic alignment. Adding CLA delivers clear gains; for example, on MDM, the R Precision@3 rises from 77.4\% to 79.1\%, and FID falls from 0.604 to 0.367. Our BiMD model achieves the best scores for both English and Chinese inputs, confirming the synergy between BiMD and CLA and setting a new performance benchmark for bilingual text-to-motion generation.

\begin{table*}[htbp]
  \centering
   \resizebox{0.95\textwidth}{!}{
        \begin{tabular}{l l l l l l l l l}
      \toprule
      \multirow{2}{*}{Method} &\multirow{2}{*}{CLA}& \multirow{2}{*}{Lang.} & \multicolumn{3}{c}{R Precision $\uparrow$} & \multirow{2}{*}{FID $\downarrow$} & \multirow{2}{*}{\makecell{MM Dist $\downarrow$}} & \multirow{2}{*}{\makecell{Div $\rightarrow$}} \\
      \cmidrule(lr){4-6}
       & & & Top 1 & Top 2 & Top 3 \\
      \midrule
      \multirow{2}{*}{Real} & \multirow{2}{*}{-} & CN & 0.543 & 0.732 & 0.821 & 0.002 & 3.338 & 10.750 \\
        & & EN &0.511 & 0.703 & 0.797 & 0.002 & 2.974 & 9.503 \\
      \cmidrule(lr){1-9}

      \multirow{4}{*}{Mo.Diffuse \cite{Zhang2024}}
       & \multirow{2}{*}{\ding{55}} & CN & 0.468$^{\pm 0.002}$ & 0.652$^{\pm 0.003}$ & 0.764$^{\pm 0.004}$ & 1.024$^{\pm 0.011}$ & 3.512$^{\pm 0.007}$ & 11.586$^{\pm 0.078}$ \\
       &  & EN & 0.476$^{\pm 0.001}$ & 0.666$^{\pm 0.004}$ & 0.775$^{\pm 0.003}$ & 0.991$^{\pm 0.028}$ & 3.356$^{\pm 0.006}$ & 9.972$^{\pm 0.076}$ \\
      \cmidrule(lr){2-9}
       & \multirow{2}{*}{\ding{51}} & CN  & 0.510$^{\pm 0.003}$ & 0.677$^{\pm 0.002}$ & 0.797$^{\pm 0.001}$ & 0.493$^{\pm 0.009}$ & 3.138$^{\pm 0.014}$ & 10.610$^{\pm 0.066}$ \\
       & & EN  & 0.508$^{\pm 0.004}$ & 0.681$^{\pm 0.003}$ & 0.802$^{\pm 0.004}$ & 0.360$^{\pm 0.006}$ & 3.131$^{\pm 0.009}$ & 9.656$^{\pm 0.056}$ \\

      \cmidrule(lr){1-9}

      \multirow{4}{*}{MDM \cite{Tevet2023}}
      & \multirow{2}{*}{\ding{55}}  & CN & 0.451$^{\pm 0.002}$ & 0.655$^{\pm 0.004}$ & 0.774$^{\pm 0.003}$ & 0.604$^{\pm 0.012}$ & 3.482$^{\pm 0.005}$ & 11.674$^{\pm 0.082}$ \\
      & & EN & 0.462$^{\pm 0.002}$ & 0.669$^{\pm 0.003}$ & 0.773$^{\pm 0.003}$ & 0.527$^{\pm 0.007}$ & 3.547$^{\pm 0.007}$ & 10.010$^{\pm 0.091}$ \\
      \cmidrule(lr){2-9}
      & \multirow{2}{*}{\ding{51}} & CN  & 0.462$^{\pm 0.003}$ & 0.671$^{\pm 0.007}$ & 0.791$^{\pm 0.004}$ & 0.367$^{\pm 0.005}$ & 3.330$^{\pm 0.007}$ & 11.122$^{\pm 0.073}$ \\
      & & EN  & 0.476$^{\pm 0.001}$ & 0.673$^{\pm 0.003}$ & 0.786$^{\pm 0.003}$ & 0.378$^{\pm 0.011}$ & 3.318$^{\pm 0.005}$ & 9.920$^{\pm 0.074}$ \\

      \cmidrule(lr){1-9}

      \multirow{4}{*}{MLD \cite{Chen2023}}
      & \multirow{2}{*}{\ding{55}}
      & CN & 0.471$^{\pm 0.005}$ & 0.668$^{\pm 0.007}$ & 0.761$^{\pm 0.007}$ & 0.743$^{\pm 0.002}$ & 3.412$^{\pm 0.003}$ & 11.214$^{\pm 0.088}$ \\
      && EN & 0.488$^{\pm 0.002}$ & 0.687$^{\pm 0.010}$ & 0.773$^{\pm 0.002}$ & 0.681$^{\pm 0.007}$ & 3.371$^{\pm 0.009}$ & 9.865$^{\pm 0.082}$ \\
      \cmidrule(lr){2-9}
      & \multirow{2}{*}{\ding{51}} & CN  & 0.504$^{\pm 0.003}$ & 0.720$^{\pm 0.008}$ & 0.793$^{\pm 0.007}$ & 0.473$^{\pm 0.005}$ & 3.196$^{\pm 0.010}$ & 10.924$^{\pm 0.079}$ \\
      & & EN & 0.515$^{\pm 0.002}$ & 0.713$^{\pm 0.007}$ & 0.807$^{\pm 0.002}$ & 0.397$^{\pm 0.005}$ & 3.105$^{\pm 0.009}$ & 9.748$^{\pm 0.065}$ \\

      \cmidrule(lr){1-9}

      \multirow{4}{*}{BiMD (Ours)}
      & \multirow{2}{*}{\ding{55}} & CN & 0.505$^{\pm 0.008}$ & 0.691$^{\pm 0.007}$ & 0.796$^{\pm 0.014}$ & 0.215$^{\pm 0.004}$ & 3.271$^{\pm 0.008}$ & 10.996$^{\pm 0.097}$ \\
      & & EN & 0.528$^{\pm 0.003}$ & 0.709$^{\pm 0.003}$ & 0.808$^{\pm 0.006}$ & 0.169$^{\pm 0.011}$ & 3.152$^{\pm 0.004}$ & 9.881$^{\pm 0.092}$ \\
      \cmidrule(lr){2-9}
      & \multirow{2}{*}{\ding{51}} & CN & \textbf{0.542}$^{\pm 0.002}$ & \textbf{0.733}$^{\pm 0.003}$ & \textbf{0.820}$^{\pm 0.003}$ & \textbf{0.060}$^{\pm 0.003}$ & \textbf{2.848}$^{\pm 0.006}$ & \textbf{10.594}$^{\pm 0.071}$ \\
      & & EN & \textbf{0.558}$^{\pm 0.003}$ & \textbf{0.731}$^{\pm 0.008}$ & \textbf{0.828}$^{\pm 0.006}$ & \textbf{0.045}$^{\pm 0.002}$ & \textbf{2.803}$^{\pm 0.005}$ &\textbf{9.619}$^{\pm 0.066}$ \\
      \bottomrule
    \end{tabular}
 }
  \caption{{Comparison of text-to-motion generation performance on the BiHumanML3D dataset.} ``Lang." indicates the evaluated language. English and Chinese results are evaluated using the T2MT evaluator~\cite{Guo2022} and our proposed evaluator, respectively. \textbf{Bold} highlights the best results.  The symbol ``\ding{51}'' at ``CLA’’ denotes methods that employ our cross-lingual alignment representation, indicating the use of a unified model for generation, whereas the symbol ``\ding{55}’’ means methods retrained with XLM as text encoder.}
  \label{tab:bih3d_result}
\end{table*}

\paragraph{Zero-Shot Text-to-Motion Generation:} To thoroughly investigate the advantages of bilingual training, we evaluate the model on zero-shot cross-lingual generation. As detailed in Table \ref{tab:zero-shot}, when trained exclusively on a single language (e.g., English) and evaluated on a completely unseen language (e.g., Chinese), for more comprehensive cross-lingual understanding, models enhanced with CLA demonstrate significantly superior performance in zero-shot cross-lingual motion generation compared to without CLA baselines. Crucially, Figure \ref{fig:zero-shot} visually confirms that our BiMD with CLA reliably generates corresponding motion from English text descriptions, despite never being trained on English data.

\begin{table}[h]
  \centering
  \resizebox{\linewidth}{!}{%
    \begin{tabular}{l c c c c c c c}
      \toprule
      \multirow{2}{*}{Text Source} & \multicolumn{3}{c}{R-Precision $\uparrow$} & \multirow{2}{*}{FID $\downarrow$} & \multirow{2}{*}{MMDist $\downarrow$} & \multirow{2}{*}{Div $\rightarrow$} \\
      \cmidrule(lr){2-4}
      & R1 & R2 & R3 & & & \\
      \midrule
      DeepL-API & 0.377 & 0.585 & 0.687 & 0.776 & 4.763 & 10.766 \\
      GPT-Translated & 0.387 & 0.593 & 0.700 & 0.721 & 4.401 & 10.396 \\
      DeepSeek-Translated & 0.393 & 0.601 & 0.713 & 0.645 & 4.253 & 10.523 \\
      HumanML3D & \textbf{0.481} & \textbf{0.673} & \textbf{0.772} & \textbf{0.473} & \textbf{3.196} & \textbf{9.724} \\
      \bottomrule
    \end{tabular}%
  }
  \caption{Performance comparison of the MLD model conditioned by translated English texts versus HumanML3D (native English texts).}
  \label{tab:cmp_translation_baseline}
  \vspace{-10pt}
\end{table}

\subsection{Ablation Studies and Discussions}

\begin{figure}[t]
    \centering
    \subfloat[OpenCLIP\label{subfig:negCL}]{
    \includegraphics[width=0.32\linewidth]{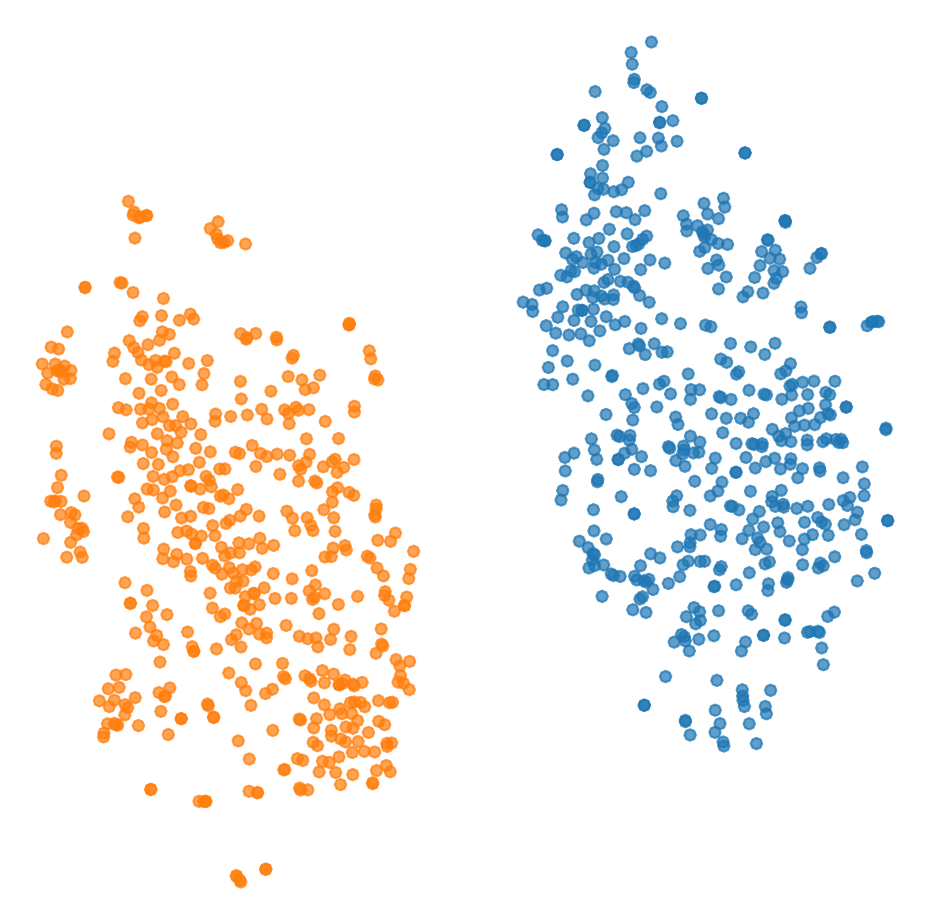}}\hfill
    \subfloat[XLM w/ CLA \label{subfig:woxc}]{
    \includegraphics[width=0.32\linewidth]{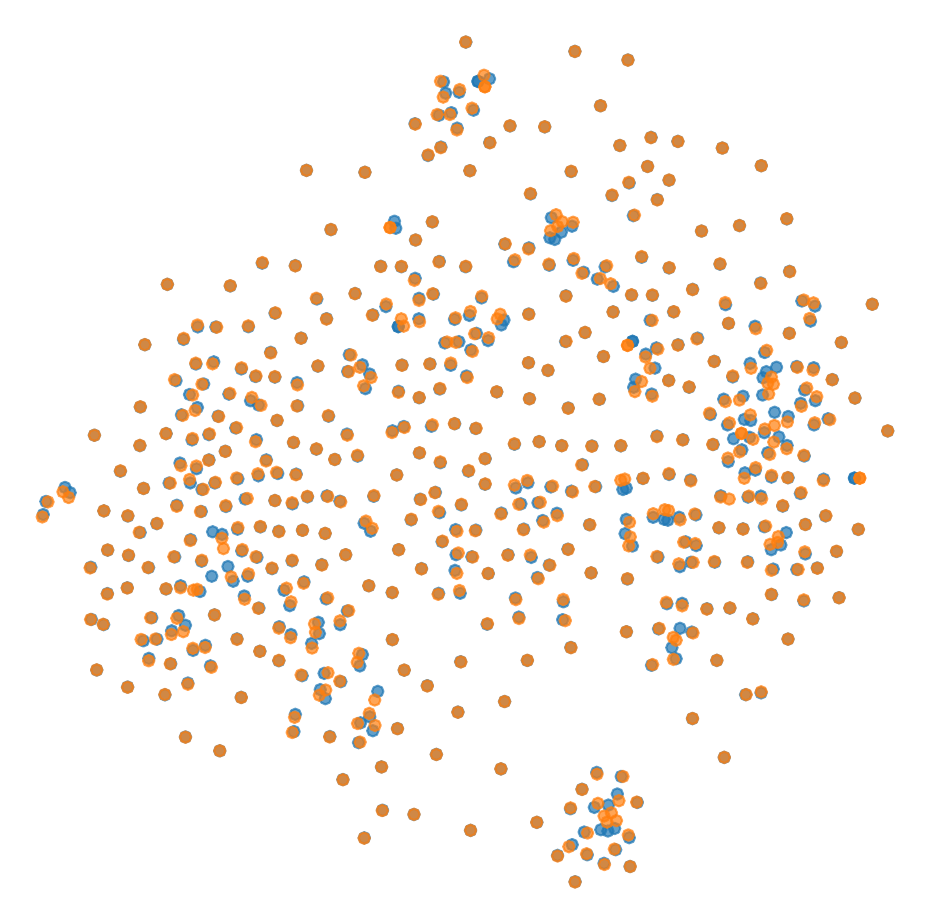}}\hfill
    \subfloat[XLM w/o CLA \label{subfig:wost}]{
    \includegraphics[width=0.32\linewidth]{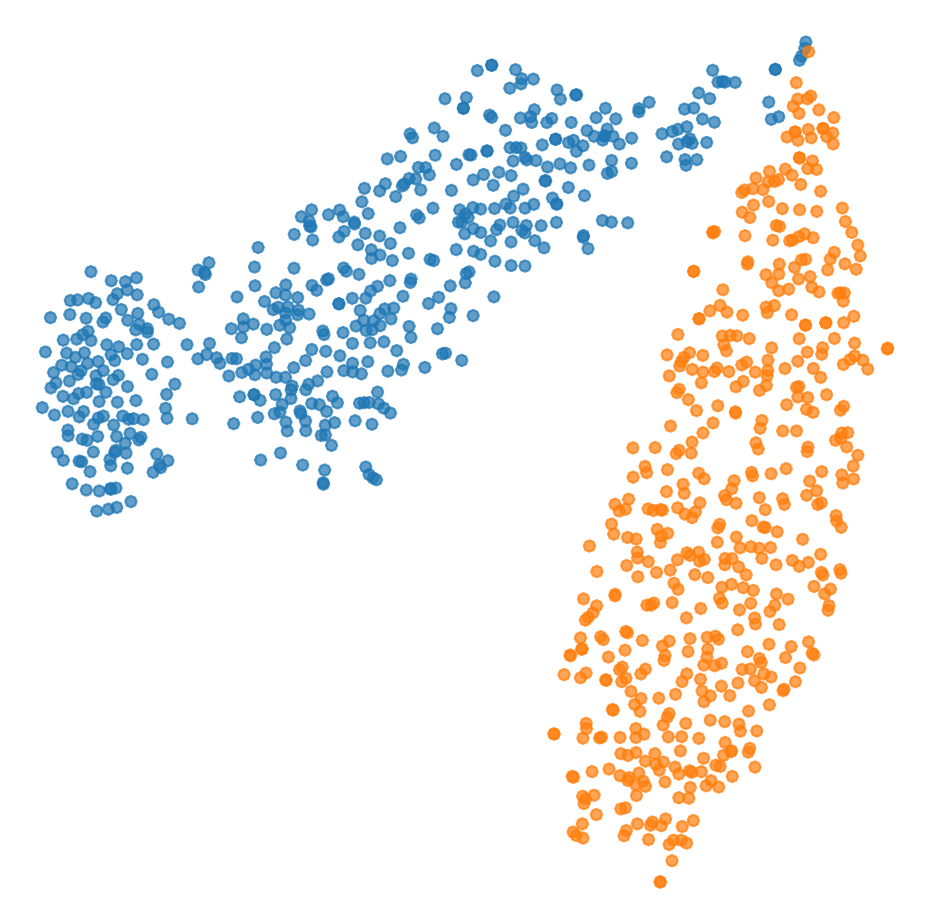}}
    \caption{t-SNE visualizations of text representations, showing the distribution of 1000 randomly selected text descriptions from the testing set of HumanML3D. Dots of the same color represent text of the same language, with potential alignment effects observed.}
    \label{fig:tsne}
\end{figure}

\paragraph{Effectiveness of Cross-Lingual Alignment:} The baseline employs the multilingual pretrained model XLM~\cite{conneau2019unsupervised}, yet insufficient cross-lingual semantic alignment leads to notable language bias, where English consistently outperforms Chinese as shown in Table \ref{tab:bih3d_result}, with further performance degradation on code-switching inputs, which is common in real-world. To address this limitation, CLA introduces cross-lingual sentence embedding alignment, as shown in Figure \ref{fig:tsne}, effectively bridging the semantic gap across languages. Quantitatively, CLA reduces the English--Chinese R@3 gap from 4.4\% (MLD w/o CLA) to 0.8\% (BiMD w/ CLA), demonstrating substantially more balanced cross-lingual performance.

\paragraph{Bilingual Training Does Not Degrade Monolingual Quality:}
A natural concern is whether joint bilingual training hurts single-language performance. Comparing BiMD w/ CLA in Table~\ref{tab:sota_humanmld3d} (monolingual setting) and Table~\ref{tab:bih3d_result} (bilingual setting), the English evaluation metrics remain identical (R@1: 0.558, R@3: 0.828, FID: 0.045), confirming that bilingual training successfully leverages shared semantics \emph{without} sacrificing language-specific quality. This addresses the question of whether a unified bilingual model offers advantages over simply training two separate monolingual models: our approach achieves identical English performance while additionally supporting Chinese, zero-shot cross-lingual transfer, and code-switching, which are capabilities that monolingual models fundamentally lack.

\begin{table}[t]
  \centering
  \small
  \setlength{\tabcolsep}{3pt}
  \renewcommand{\arraystretch}{1.15}

  \begin{tabular}{@{}cc@{}}
    \scalebox{0.85}{
      \begin{tabular}{lccc}
        \toprule
             & Zero-shot & MLD & GT \\
        \midrule
        Win Rate & 35.4      & 17.6 & 47.0 \\
        \bottomrule
        \multicolumn{4}{c}{\textbf{(a) Motion Preference (\%)}}\\
      \end{tabular}
    }%
    &
    \scalebox{0.85}{%
      \begin{tabular}{lccc}
        \toprule
             & w/ CLA & w/o CLA & Neither \\
        \midrule
        Win Rate & 68.1   & 24.8    & 7.1 \\
        \bottomrule
        \multicolumn{4}{c}{\textbf{(b) Semantic Consistency (\%)}}\\
      \end{tabular}
    }%
  \end{tabular}

      \caption{User study.
        (a) Motion preference: BiMD (Zero-shot) vs.\ MLD vs.\ ground truth on English prompts.
        (b) Semantic consistency for paired Chinese–English prompts with/without CLA. Percentages are averaged over prompts.}
  \label{tab:user_study}
\end{table}

\begin{figure}[t]
    \centering
    \includegraphics[width=1\linewidth]{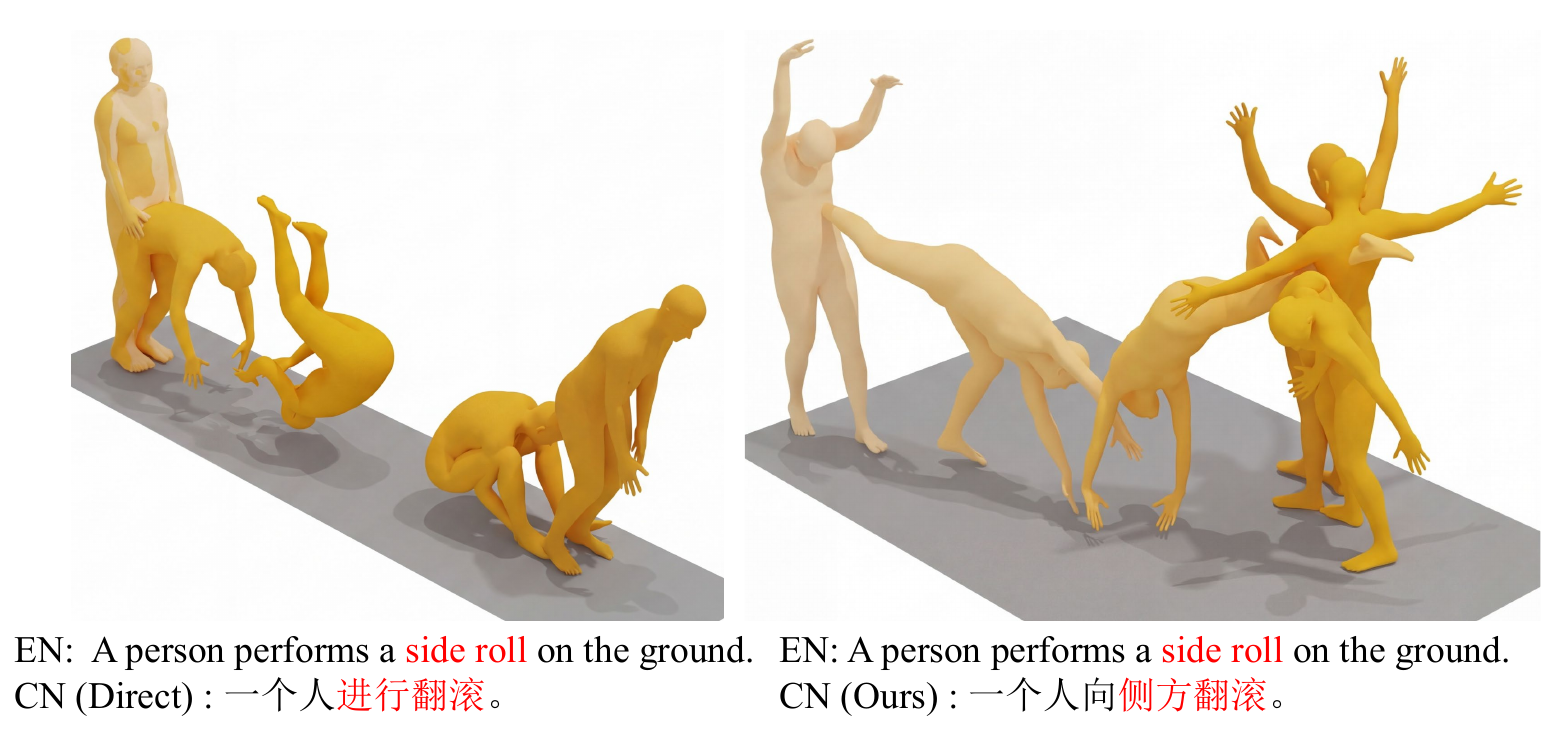}
    \caption{A translation failure case. Direct translation may omit motion-critical details (e.g., direction), resulting in under-specified descriptions. Our annotation preserves such information, leading to more precise motion semantics.}
    \label{fig:failure_translation}
\end{figure}

\paragraph{Effectiveness of LLM-assisted Annotation Pipeline:} To validate the effectiveness of our annotation pipeline and the necessity of bilingual training, we conduct comparative experiments against direct-translation baselines. Specifically, we employ DeepL-API, GPT~\cite{achiam2023gpt4}, and DeepSeek~\cite{liu2024deepseek} to translate the Chinese texts in BiHumanML3D into English and subsequently use MLD for motion generation.  Table \ref{tab:cmp_translation_baseline} demonstrates that the direct translation approaches consistently introduce significant degradation of performance. Most notably, the R Precision@3 drops from 77.2\% to nearly 70\%. We attribute this degradation to three systematic failure modes: (1) \emph{embodied-semantic distortion}, where translations preserve lexical meaning but lose body-part grounding (e.g., ``step over'' becomes generic ``walk past''); (2) \emph{temporal decomposition loss}, where multi-clause action sequences have their ordering or simultaneity altered; and (3) \emph{distribution mismatch} between translated text style and the original HumanML3D caption distribution the model was trained on. These results confirm that our curated annotation pipeline and dedicated bilingual training are essential for high-fidelity motion generation. These results confirm that our curated annotation pipeline and dedicated bilingual training are essential for high-fidelity motion generation. Besides, a qualitative example is shown in Figure~\ref{fig:failure_translation}, where direct translation omits motion-critical details (e.g., direction), leading to under-specified descriptions.

\begin{table}[htb]
  \centering
  \resizebox{\linewidth}{!}{%
    \setlength{\tabcolsep}{3pt} 
    \begin{tabular}{@{}ccccccccc@{}}
      \toprule
      \multirow{2}{*}{Train} & \multirow{2}{*}{Test} & \multirow{2}{*}{CLA} & \multicolumn{3}{c}{R Precision $\uparrow$} & \multirow{2}{*}{FID $\downarrow$} & \multirow{2}{*}{\makecell{MM Dist \\$\downarrow$}} & \multirow{2}{*}{\makecell{Div \\$\rightarrow$}} \\
      \cmidrule(lr){4-6}
      & & & Top 1 & Top 2 & Top 3 & & & \\
      \midrule
      \multirow{2}{*}{Real} & CN & \multirow{2}{*}{--} & 0.543 & 0.732 & 0.821 & 0.002 & 3.338 & 10.750 \\
      & EN & & 0.511 & 0.703 & 0.797 & 0.002 & 2.974 & 9.503 \\
      \midrule
      EN & CN & \ding{55} & 0.173 & 0.288 & 0.364 & 3.533 & 6.134 & 9.231 \\
      EN & CN & \ding{51} & \textbf{0.425} & \textbf{0.608} & \textbf{0.701} & \textbf{0.072} & \textbf{3.584} & \textbf{9.720} \\
      \midrule
      CN & EN & \ding{55} & 0.184 & 0.298 & 0.388 & 3.397 & 6.258 & 9.228 \\
      CN & EN & \ding{51} & \textbf{0.437} & \textbf{0.607} & \textbf{0.697} & \textbf{0.064} & \textbf{3.568} & \textbf{9.639} \\
      \bottomrule
    \end{tabular}%
  }
  \caption{Cross-lingual motion generation performance under different training and test language settings. Train and Test denote the training and evaluation languages, respectively. ``CLA" indicates whether Cross-lingual Alignment is applied. The arrows $\uparrow$, $\downarrow$, and $\rightarrow$ indicate higher, lower, and closer-to-real-motion values are better, respectively. \textbf{Bold} highlights the best results.}
  \label{tab:zero-shot}
\end{table}

\paragraph{User Studies:} We conduct a user study with 30 participants (15 native Chinese speakers, 15 fluent English speakers) recruited from graduate students and researchers familiar with 3D animation. Each participant evaluates 20 randomly selected text--motion pairs. For zero-shot evaluation, users compare three sets of motion videos side by side: motions generated by a model trained only on Chinese with English prompts (zero-shot), those generated by MLD, and ground truth. The presentation order is randomized and blinded. As shown in Table~\ref{tab:user_study}~(a), 35.4\% of users prefer the zero-shot results, demonstrating competitive quality despite the model never seeing English during training. For semantic consistency evaluation, we design paired Chinese and English prompts with identical semantics and generate motions using BiMD with and without CLA. Users are asked: ``Which pair of motions is more semantically consistent, or if neither was consistent?'' As shown in Table~\ref{tab:user_study}~(b), 68.1\% of users prefer the CLA version, significantly outperforming the w/o CLA variant (24.8\%), confirming that CLA substantially improves cross-lingual semantic consistency.

\section{Conclusion}\label{conclusion}
In this paper, we introduce BiHumanML3D, the first bilingual benchmark for text-to-motion generation, and propose Bilingual Motion Diffusion with Cross-Lingual Alignment. Our multi-stage annotation pipeline yields reliable Chinese-English motion pairs, while CLA explicitly bridges semantic gaps, producing balanced conditional embeddings. Extensive evaluations reveal that BiMD consistently surpasses monolingual diffusion and translation baselines, achieving accurate, culturally faithful motions even under zero-shot and code-switching scenarios. These results highlight the necessity of dedicated bilingual training and the value of semantic alignment for motion synthesis. We hope our dataset, model, and analyses catalyze broader research on human motion generation.


{
    \small
    \bibliographystyle{ieeenat_fullname}
    \bibliography{main}
}
\end{document}